\documentclass[sn-mathphys-num]{sn-jnl}

\usepackage{graphicx}%
\usepackage{multirow}%
\usepackage{amsmath,amssymb,amsfonts}%
\usepackage{amsthm}%
\usepackage{mathrsfs}%
\usepackage[title]{appendix}%
\usepackage{xcolor}%
\usepackage{textcomp}%
\usepackage{manyfoot}%
\usepackage{booktabs}%
\usepackage{algorithm}%
\usepackage{algorithmicx}%
\usepackage{algpseudocode}%
\usepackage{listings}%
\usepackage{color}
\usepackage{changes}
\usepackage{float}                  
\usepackage{subfig}        
\usepackage{overpic} 
\usepackage{graphicx}
\usepackage{multirow}

\theoremstyle{thmstyleone}%
\theoremstyle{thmstyletwo}%

\theoremstyle{thmstylethree}%

\begin{document}

\title[]{Discrete Wavelet Transform-Based Capsule Network 
for Hyperspectral Image Classification}


\author[1]{\fnm{Zhiqiang} \sur{Gao}}\email{zgao@wku.edu.cn}

\author[2]{\fnm{Jiaqi} \sur{Wang}}\email{202213045115025@hainnu.edu.cn}

\author[3]{\fnm{Hangchi} \sur{Shen}}\email{stephen1998@email.swu.edu.cn}

\author[4]{\fnm{Zhihao} \sur{Dou}}\email{zd55@duke.edu}

\author[1]{\fnm{Xiangbo} \sur{Zhang}}\email{1307887@wku.edu.cn}

\author[5]{\fnm{Kaizhu} \sur{Huang}}\email{Kaizhu.Huang@dukekunshan.edu.cn}







\affil[1]{\orgdiv{College of Science, Mathematics and Technology}, \orgname{Wenzhou-Kean University}, \orgaddress{\street{88 Daxue Road}, \city{Wenzhou}, \postcode{325060}, \state{Zhejiang Province}, \country{China}}}

\affil[2]{\orgdiv{College of Primary Education}, \orgname{Hainan Normal University}, \orgaddress{\street{No. 99 Longkun South Road}, \city{Haikou}, \postcode{571158}, \state{Hainan Province}, \country{China}}}

\affil[3]{\orgdiv{College of Artificial Intelligence}, \orgname{Southwest University}, \orgaddress{\street{No. 2 Tiansheng Road}, \city{Chongqing}, \postcode{400715}, \state{Sichuan Province}, \country{China}}}

\affil[4]{\orgdiv{Pratt School of Engineering}, \orgname{Duke University}, \orgaddress{\street{305 Teer Engineering Building}, \city{Durham}, \postcode{27708}, \state{North Carolina}, \country{USA}}}

\affil[5]{\orgdiv{Data Science Research Center \& Division of Natural and Applied Sciences}, \orgname{Duke Kunshan University}, \orgaddress{\street{No. 8 Duke Avenue}, \city{Suzhou}, \postcode{215316}, \state{Jiangsu Province}, \country{China}}}


\abstract{

Hyperspectral image (HSI) classification is a crucial technique for remote sensing to build large-scale earth monitoring systems. HSI contains much more information than traditional visual images for identifying the categories of land covers. One recent feasible solution for HSI is to leverage CapsNets for capturing spectral-spatial information. However, these methods require high computational requirements due to the full connection architecture between stacked capsule layers. 
To solve this problem, a DWT-CapsNet is proposed to identify partial but important connections in CapsNet for a effective and efficient HSI classification. Specifically, we integrate a tailored attention mechanism into a Discrete Wavelet Transform (DWT)-based downsampling layer, alleviating the information loss problem of conventional downsampling operation in feature extractors. Moreover, we propose a novel multi-scale routing algorithm that prunes a large proportion of connections in CapsNet. 
A capsule pyramid fusion mechanism is designed to aggregate the spectral-spatial relationships in multiple levels of granularity, and then a self-attention mechanism is further conducted in a partially and locally connected architecture to emphasize the meaningful relationships.
As shown in the experimental results, our method achieves state-of-the-art accuracy while keeping lower computational demand regarding running time, flops, and the number of parameters, rendering it an appealing choice for practical implementation in HSI classification.
}




\maketitle
·
\section{Introduction}\label{sec1}


Hyperspectral image (HSI) classification is a crucial technique to build large-scale earth monitoring systems.
The land-cover information is collected by hyperspectral imaging sensors (known as spectrometers) equipped on satellites and presented in the form of HSIs. Each HSI includes hundreds of narrow, contiguous spectral bands (channels) to record electromagnetic spectrum in the range from visible to short-wave infrared wavelengths over large observed areas \cite{rustamov2018multi, stuart2019hyperspectral}. Therefore, HSI contains much more information than traditional visual images, so they can be well used to distinguish the categories of land covers to achieve a wide range of earth monitoring applications, such as ecological science, precision agriculture, and surveillance services.

Previously, various robust and accurate classification schemes have been widely investigated, ranging from the classic machine learning methods (e.g., support vector machines \cite{camps2005kernel}, k-means clustering \cite{haut2017cloud}, Gaussian process \cite{bazi2009gaussian}, random forest (RF) \cite{ham2005investigation}, and extreme learning machines \cite{haut2018fast}) to the deep neural network classifiers (such as stacked autoencoders \cite{chen2014deep, li2023deep}, sparse autoencoders \cite{zhao2017spectral}, and deep belief networks \cite{li2014classification}).
Among them, approaches based on Convolutional Neural Networks and Capsule Networks are widely investigated and achieve the excellent performances.

\subsection{Convolutional Neural Networks for HSI Classification}

Convolutional Neural Networks (CNNs) have demonstrated excellent performances in processing the spectral-spatial information contained in HSI for classification. 
One advantage of CNN is automatically extracting the hierarchical features of an object, where the features will be refined layer by layer until the final abstract features meet the requirement of classification \cite{chen2022integrated}.
Notably, the CNN extracts representative features of the object regardless of its position in space (translation invariance).
This is achieved by a kernel mechanism employing the locally connected and shared weights to detect whether the meaningful features are present or not, followed by a downsampling operation, such as the average or max pooling method.

The inherent architectures of CNNs have been widely studied to learn the spectral-spatial information in HSI. To fully take advantage of the spectral signature (each pixel vector on HSI), the 1D CNN model \cite{chen2016deep, paoletti2018new, haut2018active} is leveraged for feature extraction, where CNNs work in a pixel-wise manner and take each spectral vector as input. However, the spectral information at each pixel is highly mixed, which inevitably introduces intra-class variability and inter-class similarity into HSI.
As such, a 2D CNN model, taking a 2D HSI patch as input, is leveraged for feature extraction, which is motivated by the straightforward assumption that adjacent pixels are correlated and belong to the same category \cite{gao2018convolution}.
Moreover, combining the spectral signature and spatial patch information, the 3D CNNs \cite{rs14215334, 9526708} naturally reduce the intra-class variability and present superior performances.

Due to the high complexity and rich spectral-spatial information of HSI data, it is natural to adapt deeper DNNs to HSI classification.
However, this practice will lead to some potential problems, such as gradient vanishing and the parameter volume exploding.
In detail, as the number of DNN layers increases, the gradient of its parameters will gradually vanish during the training process, which will make it difficult for network parameters to be fully trained and lead to a slow convergence speed \cite{srivastava2015training}.
To solve this problem in HSI classification, more advanced architectures, such as residual connection \cite{he2016deep, zhong2017spectral, paoletti2018deep} (ResNet) and dense skip connection \cite{huang2017densely, paoletti2018deep} (DenseNet), were introduced to discover richer spectral-spatial features.
Meanwhile, as the amount of parameters increases, a large amount of data samples needs to be added to the training processing to prevent overfitting problems of DNN. One possible solution is to leverage data augmentation mechanism, such as increasing the diversity of training samples \cite{chen2016deep, acquarelli2018spectral} or generating new samples for minority classes \cite{roy2021generative}, to certify better generalization capabilities of DNNs.
Additionally, CNN models are also known for their resource-intensive nature. They demand substantial computational power and storage capacity due to their large number of parameters. This can pose challenges in terms of feasibility for deployment, particularly in scenarios with limited computational resources.

More importantly, CNNs present poor robustness when there is a noticeable change in the object's orientation or pose 
because they cannot identify the object's geometric transformations (such as rotations and shift movements \cite{paoletti2020rotation}) and discard spatial relationships among features, such as size and perspective \cite{sabour2017dynamic}. 
Actually, the max-pooling operation of CNNs only focuses on translation invariance by detecting whether entities' features exist regardless of where they are in space, which inevitably loses useful spatial information.
Some approaches have tried to encode the transformation invariances and symmetries that exist in the
data into features by using the data augmentation methods, which conduct predefined transformations on original data samples during training  \cite{lecun2010convolutional, zhao2016spectral}.
However, the spatial relationships among features have yet to be fully explored in these approaches.

\subsection{Capsule Networks for HSI Classification}

To overcome the information loss of spatial relationships in the downsampling operation of CNNs,
Capsule Networks (CapsNet) are introduced as an innovative approach to enhance the representational capabilities of neural networks \cite{sabour2017dynamic}. 
CapsNet employs multi-layer capsules and a routing algorithm (routing-by-agreement) to encode part-whole relationships between different entities (features, objects, or object parts) into the instantiation/activity vectors (capsules), where captured relationships may include pose, orientation, scale, illumination, and so on.
Following the guidance of the routing algorithm, all capsules in a low-level layer iteratively interact with one of the capsules in a high level and then are aggregated to update capsules in a high-level layer by using the voting mechanism.
Such a ``full connection" interaction process between capsule layers actively encodes the spatial relationships of entities.
In the final classification, the direction of the activity vector represents the properties of the object, and the length indicates the probability of belonging to a class.

For HSI classification, CapsNet has also been proven to be an effective learner for capturing spatial relationships among spectral features in a pioneered study \cite{paoletti2018capsule}.
Some following studies focus on investigating novel CapsNet structures for enriching spatial relationships among capsules. 
Jiang et al. \cite{jiang2020spectral} presented a dual-channel CapsNet, where features from spectral and spatial domains are extracted by two separate convolution channels separately and then are concatenated together for downstream calculations in capsule layers.
Zhu et al. \cite{zhu2019deep} propose the 1D-CapsNet that extracts the spatial and spectral features by using a capsule-wise constraint window method to decrease the number of parameters.
Additionally, the proposed routing algorithm leverages the shared weights between high-level and low-level capsules.
Nevertheless, simply sharing weights has limitations in generating a representative high-level capsule, which makes capsules in two adjacent layers similar. 
Therefore, directly sharing weights in the routing algorithm will constrain model performance and cause terrible results to some extent.

On the other hand, some studies are dedicated to enriching spectral-spatial information in feature maps in order to promote spatial relationships captured by CapsNet. 
Li et al. \cite{li2020robust} introduce the maximum correntropy criterion (MCC) to control the negative impact introduced by the noise and outliers. Meanwhile, the MCC-based dual-channel CapsNet framework is designed for fusing hyperspectral data and LiDAR data to gain powerful feature maps. 
Paoletti et al. \cite{9650856} propose a multiple attention-guided CapsNet and apply the various attention methods to input data, feature map, and capsule routing, where multiple attention mechanisms cooperatively extract the more representative visual parts of the images for HSI classification. 
Lei et al. \cite{lei2022multiscale} design a multi-scale feature aggregation CapsNet, which consists of two branches to extract features from local spatial images and global spatial images. The multi-scale global and regional feature maps are combined to improve model performance.
Lei et al. \cite{lei2021hyperspectral} investigate a 3D Convolutional Capsule layer and a novel dynamic routing that connects the adjacent capsules based on 3D convolution to capture more robust higher-level feature information with less number of parameters.

\subsection{Contributions}


Although CapsNet has achieved great success in HSI classification, 
both training and reasoning with CapsNet are computationally expensive, which limits its wide deployment in real-world application.
The high computation complexity of CapsNet is mainly attributed to the full connection between the capsule layers in the routing algorithm.
Previous studies have contributed wisdom to overcome this drawback by designing efficient routing algorithms, such as replacing the original routing method with an Expectation-maximization algorithm \cite{hintonMatrixCapsulesEM2018}, multiple experts ensembling mechanism \cite{hahnSelfRoutingCapsuleNetworks2019}, or combination of Gumble sampling and attention mechanism \cite{ahmedSTARCapsCapsuleNetworks}.
On the other hand, one practical solution is to utilize a CNN-CapsNet strategy, where a simple CNN \cite{rajasegaranDeepCapsGoingDeeper2019} or ResNet \cite{hahnSelfRoutingCapsuleNetworks2019} is employed as the backbone for feature extraction, and then forward pass the latent features to a few capsule layers to learning spatial relationship among feature patches.
This strategy is helpful to adapt CapsNet to more complex data, such as HSI, and avoid stacking more capsule layers.
However, the improvements brought by these methods are limited, and the computational efficiency still needs to be further improved.

To enable a more efficient CNN-CapsNet strategy for HSI classification, we aim to prune the full connection between capsule layers in the routing algorithm.
However, pruning some connections directly will also lead to further information loss of spatial relationships, resulting in a trade-off between accuracy and efficiency.
Moreover, since conventional max or average pooling operations still exist in the CNN-based backbones, there is a large amount of information loss for spectral-spatial information before executing the routing algorithm.
As such, if the above information loss can be alleviated effectively in the downsampling layer, pruning some connections can still preserve the learning ability for spatial relationships and reduce the computational complexity simultaneously.

From the above statement, we propose a novel Discrete Wavelet Transform-based CapsNet (DWT-CapsNet) architecture.
In detail, we integrate a tailored attention mechanism into a DWT-based downsampling layer \cite{williams2018wavelet, lv2023hybrid}, which enables our DWT-CapsNet to preserve rich frequency information for learning spatial relationships. 
Moreover, a multi-scale routing mechanism is proposed to enable a more effective and efficient CapsNet.
Leveraging the insight from Feature Pyramid Network \cite{lin2017feature}, we design a Pyramid Fusion approach that aggregates the spatial relationships in multiple levels of granularity to further enrich spatial relationship information for the downstream calculations.
Building upon the above components that are designed for information preservation,
we investigate a partially and locally connected structure for high-level capsule generation.
Additionally, we further explore the partial but important connections by using a sliding window and self-attention mechanism.
As such, compared with vanilla CapsNet employing the fully connected architecture, 
the proposed method is able to reduce the computational demand without catastrophic information loss.
As shown in the experimental results, 
our method achieves state-of-the-art accuracy while keeping lower computational requirements compared with previous CapsNets for HSI classification.

The following are the contributions of this paper: 


(1) 
We design a novel attentive DWT downsampling layer to alleviate the information loss caused by max-pooling operations in CNNs.
Our downsampling method integrates an attention mechanism that is jointly trained with CapsNets to learn applicable frequency ranges for capturing spatial relationships among features.

(2) 
Different from vanilla dense routing methods \cite{ahmedSTARCapsCapsuleNetworks,hahnSelfRoutingCapsuleNetworks2019} in the spatial domain, our novel routing method conducts a Pyramid Fusion on the frequency domain, where multi-scale entities are aggregated to capture scale invariant spatial relationships.
Meanwhile, with a partially connected architecture between capsule layers, our method not only reduces the computational demand but also retains more helpful information related to the task.

(3) As shown in the results and analysis,
our method achieves state-of-the-art accuracy while keeping lower computational demand regarding running time, flops, and the number of parameters, rendering it an appealing choice for practical implementation in HSI scenarios.

\section{Method}

In this section, we show the details of the proposed DWT-CapsNet. 
Firstly, we introduce the overview architecture of our method, and the corresponding data pipeline.
Then, we briefly describe the fundamentals of CapsNet.
Finally, our novel architectures, including the attentive DWT downsampling filter layer and multi-scale routing method, are illustrated in detail.

\subsection{Overview}

\begin{figure}
    \centering
    \resizebox{0.9\textwidth}{!}{\includegraphics{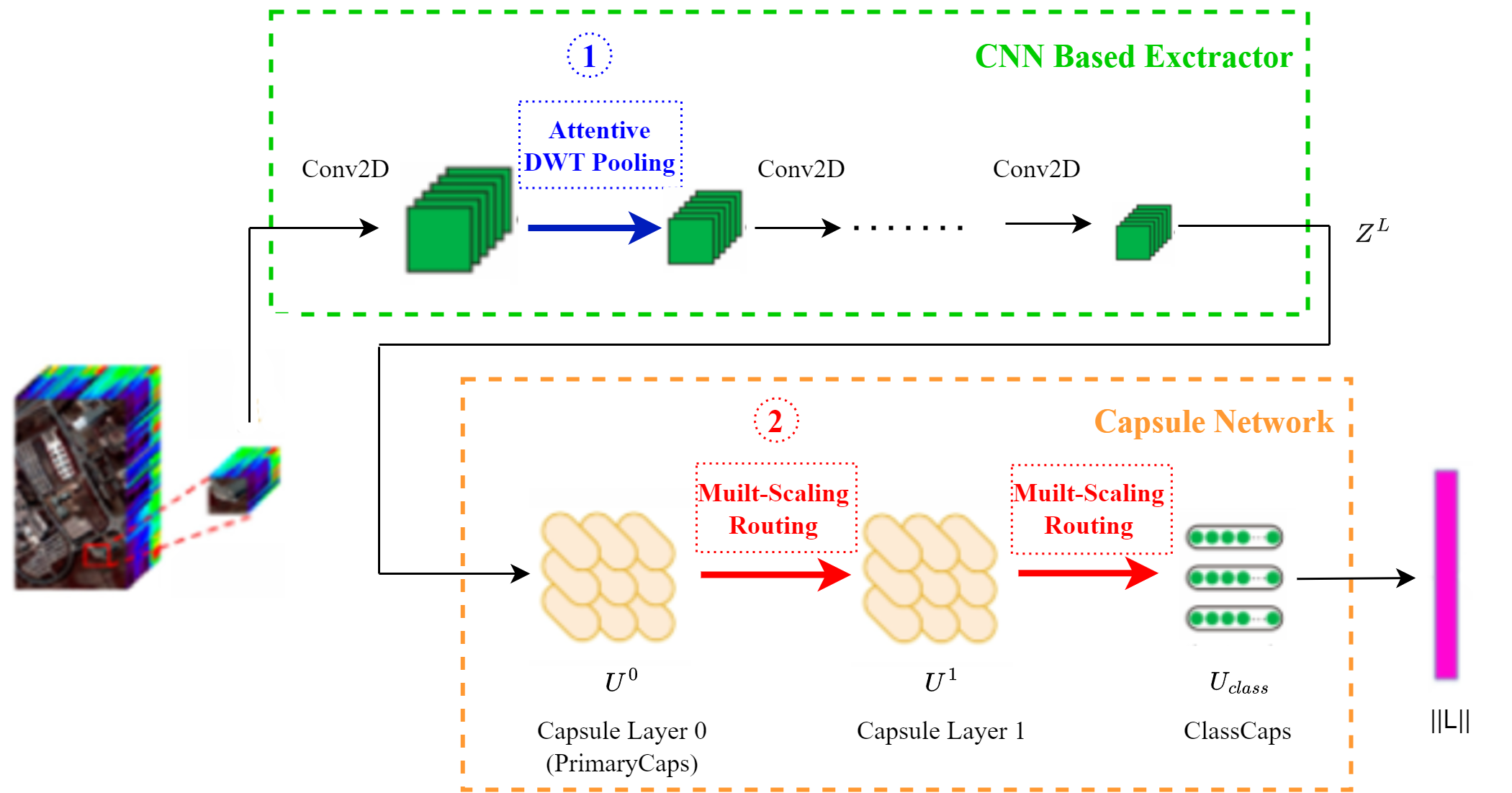}}
    \caption{Overall DWT-CapsNet processing:
1. The CNN backbone will extract the feature maps from the HSI patches, and the attentive DWT poolings have been replaced with the traditional downsampling methods, which are embedded into the CNN backbone.
2. The extracted feature map is then input into PrimaryCaps and transferred into capsule structure in layer $U^1$. The capsule network is composed of three capsule layers which are connected by multi-scale routing processes.
3. There is the number of class capsules in ClassCaps, and the maximum length of capsule $||L||$ represents the highest probability of the corresponding class.}
    \label{overview}
\end{figure}

To adapt the CapsNet to complex HSI data and reduce the computation demand, we employ the CNN-CapsNet strategy, as shown in Figure~\ref{overview}.
Firstly, the HSI patch $x \in \mathbb{R}^{n \times n \times k}$ is inputted into a CNN backbone, where the generated feature maps in the final ($L$-th) CNN layer is denoted as $Z^{L}\in \mathbb{R}^{ N \times N \times K}$.
Then, $Z^{L}$ is forward passed through the CapNet to further explore the spectral-spatial information in HSI for classification, where the PrimaryCaps transfers the feature maps into activity tensors $U^{0} \in \mathbb{R}^{N_{c} \times N_{c} \times 2p}$.

The proposed two components DWT-based downsampling and Multi-scaling Routing method are embedded in CNN backbone and CapNet respectively.
Specifically, our DWT-based downsampling will replace all the conventional downsampling operations in the CNN backbone to alleviate the spectral-spatial information loss. 
Our Multi-scaling Routing method, located between two capsule layers, generates high-level capsules for capturing spectral-spatial relationships.
Finally, the ClassCaps will flatten all activity tensors to vectors and classify the HSI data based on the maximum length of these vectors, which will be introduced in detail in Section~\ref{sec:backgroud_capsnet}.

\subsection{Background of Capsule Network}\label{sec:backgroud_capsnet}

To capture the part-whole relationships among entities (target object or object’s part of interest), the CapsNet employs multiple-layer capsules, and capsules between layers are forced to interact by implementing routing algorithms. 
This design organizes the capsules in a tree-like architecture, where each node of the tree corresponds to an active capsule. 
Moreover, in the forward passing stage, each capture in low layers votes for a capsule state in a high layer (parent), and parents aggregate the votes and update their states, which built a voting mechanism for capturing spatial relationships.

Specifically, the feature maps, $Z^{L} \in R^{N \times N \times K}$ produced by backbone (CNNs) in the last convolutional layer (e.g., $L$-th layer), are reshaped into a tensor of $N \times N \times\left(K/S \times \mathbb{R}^S\right)$ low-level entities in PrimaryCaps,
where $S$ is the number of neurons inside each capsule and $K/S$ is the number of capsules.
The capsules in PrimaryCaps work as the initial ones to be involved in a layer-by-layer routing algorithm.
The output for $i$-th capsule from $l$-th capsule layer is denoted as $u_i^l$ which captures the spatial information among features.
For each capsule, given a corresponding input $s_i^l$, $u_i^l$ is obtained by a nonlinear squashing function,
\begin{equation}
    u_j^{l} = \frac{||s_{j}^{l}||^2}{1+||s_{j}^{l}||^2} \frac{s_{j}^{l}}{||s_{j}^{l}||}.
    \label{squash}
\end{equation}
This operation forces the length of $u_i^l$ to fall in the range of 0 to 1 and to represent the probability that the meaningful feature exists or not.
Meanwhile, $u_i^l$ acts as an instantiation/activity vector, where its elements are considered its instantiation parameters (properties) and determine the pose and orientation of the feature.

When the outputs of $(l-1)$ capsule layers $u_i^{l-1}$ are given, the process used for generating $s_j^l$ relies on the routing-by-agreement mechanism.
In detail, the prediction vectors $u_{j \mid i}^{l}$ are generated
by conducting a linear transformation on $u_i^l$ firstly,
\begin{equation}{}
	\hat{u}_{j \mid i}^l ={w}_{i,j}^{l} \cdot u_{i}^{l-1}, \
 \label{eq:prediction_vectors}
\end{equation}
where ${w}_{i,j}^{l}$ represents the weight matrices.
Each $\hat{u}_{j \mid i}^l$ represents the basis for seeking the parents of the $(l-1)$-th layer capsules, i.e., those high-level capsules to which the outputs of the low-level capsules will be routed.

To conduct a parent search by agreement between high and low-level capsules, 
the $s_i^l$ is calculated by a weighted sum of $\hat{u}_{j \mid i}^l$, denoted as:
\begin{equation}
s_j^l =\sum_i c_{i j}^l \hat{u}_{j \mid i}^l \quad \text { where } c_{i j}^l=\frac{\exp \left(b_{i j}\right)}{\sum_k \exp \left(b_{i k}\right)},
\label{fulls}
\end{equation}
where $c_{i j}^l$ determined the importance of capsules $u_{i}^{l}$ occupied in the high-level capsule $u_j^{l}$ and $\sum_{i=1} c_{i,j}^{l}=1$.
Meanwhile, these $b_{i k}$ act as the $\log$ priors which denote the probability of the $i$-th low-level capsule activating the $k$-th high-level capsule, capturing the relationship between both capsules.
Indeed, this relation evolves as the routing procedure iteratively, so each $b_{i k}$ is initialized to 0, and then, it is updated according to the agreement between the prediction made and the obtained output, i.e., $<u_{k}^{l}, \hat{u}_{k \mid i}^{l}>$ as:
\begin{equation}
b_{ik}[t] =  b_{ik}[t-1] + <u_{k}^{l}, \hat{u}_{k \mid i}^{l}> [t-1],
\label{repeat}
\end{equation}
where $t$ and $t-1$ indicate the current and previous iterations of the routing procedure.


The last layers for classification are denoted as ClassCaps $U_{class}$:
\begin{equation}
    U_{class} = \{u_1^{L_{c}}, ..., u_c^{L_{c}}, ..., u_C^{L_{c}}\},
\end{equation}
where the subscript $C$ of $u$ is the number of classes in the dataset.
%
The maximum length $||L||$ among capsules is shown as:
\begin{equation}
    ||L|| = max(\{||u_{1}^{L_{c}}||, ...,, ||u_{c}^{L_{c}}||, ...,,||u_{C}^{L_{c}}||\}),
\end{equation}
which indicates the highest probability that the image belongs to which class,
i.e., the $||u_c^{L_{c}}||$ are the maximum, and the images belong to $c$-th class.

The loss function for training CapsNet for $c$-th class is calculated by:
\begin{equation}
    L_{\text {margin }}=\sum_{c}^{C}\left(T_{c} \max \left(0, \alpha^{+}-\left\| u_c^{L_{c}} \right\|\right)^{2}\right.  \\
 \left. +\lambda\left(1-T_{c}\right) \max \left(0,\left\| u_c^{L_{c}} \right\|-\alpha^{-}\right)^{2}\right),
\label{op:loss}
\end{equation}
where the value of $T_c$ is $1$ when the data belongs to the $c$-th class and is $0$ otherwise. 
Additionally, a regularization parameter denoted as $\lambda=0.5$ is introduced. Furthermore, two threshold values, $\alpha^+$ and $\alpha^-$, are defined to impose constraints on the magnitude of the capsule's activity vector $u_c^{L_{c}}$. Specifically, when $T_c=1$, the goal is to keep $\|u_c^{L_{c}}\|$ within the range $[0.9, 1]$, while for $T_c=0$, the aim is to restrict it to $[0, 0.1]$. 

Additionally, the CapsNet can be viewed as an encoder network. In this context, a decoder (e.g., a Multi-Layer Perceptron) can be incorporated to handle data reconstruction, where the final layer's outputs are used to reconstruct the input data of the whole network, such as HSI data.
Consequently, the reconstruction loss ($L_{\text{recon}}$) is calculated as the norm of the difference between the original data and the reconstructed data. This reconstruction loss can be combined with Equation \ref{op:loss} to improve the overall classification performance, resulting in the final loss ($L_{\text{final}}$) as the sum of margin loss ($L_{\text{margin}}$) and reconstruction loss ($L_{\text{recon}}$).

\subsection{Attentive Discrete Wavelet Transform Downsampling}

To circumvent the indiscriminate removal of both redundant and remaining essential information in features, 
several researchers advocate for an effective downsampling approach to improve CNN performance on general image classification tasks \cite{yu2014mixed,zeiler2013stochastic,williams2018wavelet,zhou1988computation}.
Wavelet pooling \cite{williams2018wavelet} demonstrates that transforming the feature representations in the frequency domain is a feasible approach for general classification, where the 1D DWT is applied twice on each layer, and then the high-frequency content is roughly removed for downsampling operation. 
Meanwhile, recent studies for learning models in the frequency domain \cite{xuLearningFrequencyDomain2020} also show that CNNs are indeed biased toward low-frequency content of images, where frequency transformation operations are only executed prior to the forward passing in the backbone.
%
On the other hand, 3D-Discrete Wavelet Transform is also leveraged to transform the raw HSI, and then combined with classical machine learning methods for classification \cite{rs13071255}, such as random forest, K-nearest neighbor, and support vector machine (SVM).

Obviously, the above design inevitably neglects to leverage high-frequency information for classification.
To explore more information, our downsampling layer customizes an attention method to emphasize important frequency domains for classification, where features that include low- and high-frequency information are downsampled by weighed sum operation.

\begin{figure}
    \centering
    \resizebox{1.0\textwidth}{!}{\includegraphics{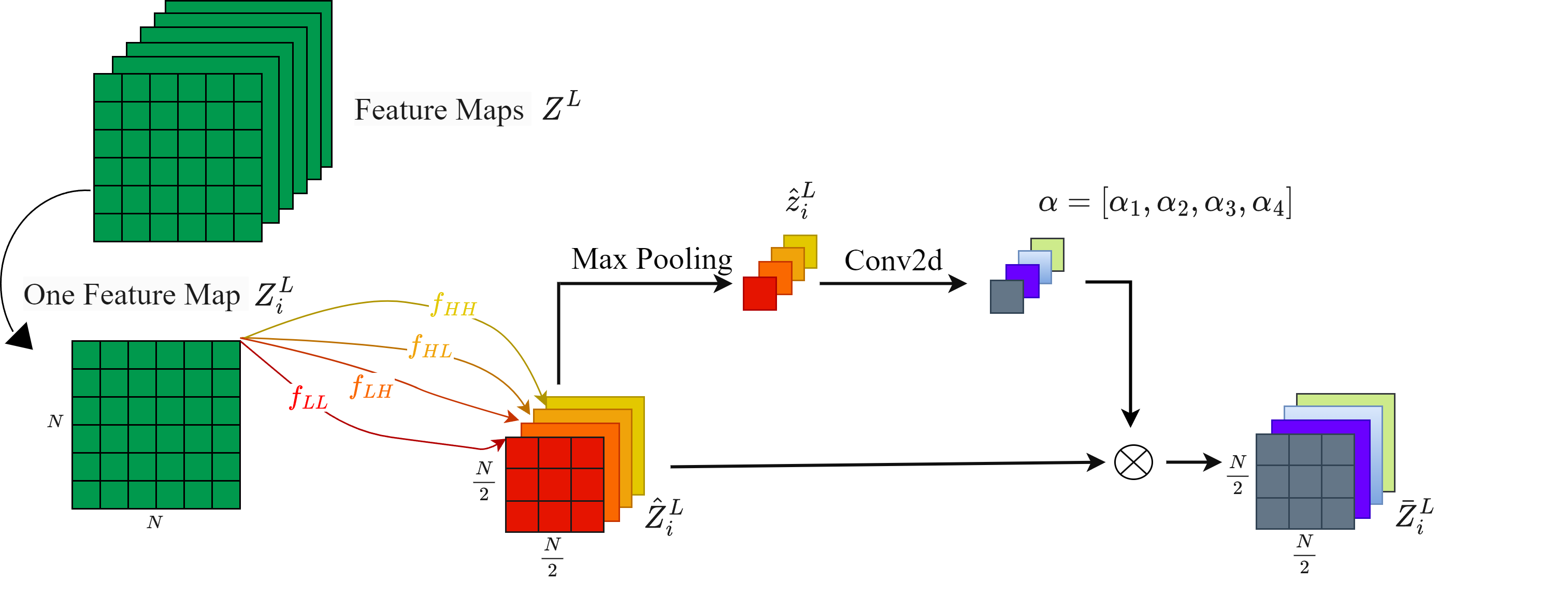}}
    \caption{
    Illustration for Attentive DWT downsampling method. 
    The DWT transfers the feature map into four channels by using filters $f_{LL},f_{LH},f_{HL}$, and $f_{HH}$. 
    Then, each channel will obtain a weight for generating the downsampled feature maps.}
    \label{dwt}
\end{figure}

Let the 3-dimensional tensor $Z^{L} \in \mathbb{R}^{N \times N \times K}$ be feature maps of a data sample in $L$-th layer of a CNN-based backbone, where $N$ is the size, and  $K$ is the number of channels. 
As shown in Figure~\ref{dwt}, taking $i$-th feature map $Z^{L}_{i} \in \mathbb{R}^{N \times N}$ as an example, after applying a 2D-DWT function $f_{DWT}(\cdot)$ \cite{mallat1989theory}, the obtained feature maps are denoted as:
\begin{equation}
\hat{Z}^{L}_{i} = f_{DWT}\left( Z^{L}_{i} \right) \in \mathbb{R}^{\frac{N}{2} \times \frac{N}{2} \times 4},
\label{eq3.1}
\end{equation}
where $4$ represents the number of frequency domains.
Specifically, there are four 2D-DWT convolutional ﬁlters, including low-pass and high-pass ﬁlters denoted as:
\[
\begin{array}{cc}
f_{LL} = \begin{bmatrix}
1 & 1 \\
1 & 1
\end{bmatrix}, &
f_{LH} = \begin{bmatrix}
-1 & -1 \\
1 & 1
\end{bmatrix}, \\
f_{HL} = \begin{bmatrix}
-1 & 1 \\
-1 & 1
\end{bmatrix}, &
f_{HH} = \begin{bmatrix}
1 & -1 \\
-1 & 1
\end{bmatrix},
\label{trans}
\end{array}
\]
which are mutually orthogonal and collectively constitute a $4 \times 4$ invertible matrix. 
These filters decompose the original feature map $Z^{L}_{i}$ into corresponding sub-band parts, which is shown as:
\begin{equation}
\centering
\begin{aligned}
\hat{Z}^{L}_{i, LL}=\left(\boldsymbol{f}_{L L} * Z^{L}_{i} \right) \downarrow_{2}, \quad& \hat{Z}^{L}_{i, LH} =\left(\boldsymbol{f}_{L H} * Z^{L}_{i} \right) \downarrow_{2},\\
\hat{Z}^{L}_{i, HL}=\left(\boldsymbol{f}_{H L} * Z^{L}_{i} \right) \downarrow_{2}, \quad & \hat{Z}^{L}_{i, HH} =\left(\boldsymbol{f}_{H H} * Z^{L}_{i} \right) \downarrow_{2},    
\end{aligned}
\end{equation}
where  $*$  represents the convolution operator, and  $\downarrow$  denotes the downsampling with stride $2$.
As a result, the obtained $\hat{Z}^{L}_{i}$ is composed of a sequence of feature maps of four different domains, where $\hat{Z}^{L}_{i} = [\hat{Z}^{L}_{i, LL}, \hat{Z}^{L}_{i, LH}, \hat{Z}^{L}_{i, HL}, \hat{Z}^{L}_{i, HH}]$ and  $\hat{Z}^{L}_{i, LL}, \hat{Z}^{L}_{i, LH}, \hat{Z}^{L}_{i, HL}, \hat{Z}^{L}_{i, HH} \in \mathbb{R}^{\frac{N}{2} \times \frac{N}{2}}$.
Notably, applying these 2D-DWT downsamplings allows a modified model to maintain almost the same number of parameters and computational complexity as the original model.

To allow the model to automatically identify the meaningful information in different frequency domains, the feature maps in $\hat{Z}^{L}_{i}$ will multiply a sequence of weights produced by our attention mechanism.
Firstly, we apply a global maxing pooling \cite{add1} on each feature map in $\hat{Z}^{L}_{i}$, where the maximum scalar values in different domains are selected for further processing.
Thus, $\hat{Z}^{L}_{i}$ is transformed to a feature vector $\hat{z}^{L}_{i} \in \mathbb{R}^{4}$.
Then, a Conv2D operation (with the kernel size $1 \times 1$) is applied on $\hat{z}^{L}_{i}$ to get a weight set $\alpha=[\alpha_1,\alpha_2, \alpha_3,\alpha_4] \in \mathbb{R}^{4}$, and each weight represents the importance of each frequency channel. 
Finally, the obtained downsampled feature maps in our downsampling operation are calculated by
\begin{equation}
    \bar{Z}^{L}_{i} = \hat{Z}^{L}_{i} \times \alpha.
\end{equation}
As such, compared with the conventional pooling method, more useful information for classification is preserved and emphasized in the obtained $\bar{Z}^{L}_{i}$ that will be used in the downstream layers.

\subsection{Muilt-Scale Routing Method}

The proposed routing method includes two components that will be introduced separately in this section.
Firstly, the Pyramid Fusion method aims to organize the spectral-spatial information preserved in capsules in a multi-scale manner. 
Then, in the Partial Connection method,  
a partial connection operation aggregates the capsules based on their activated values to update the capsules of the high-level layer, where the activated values are derived by using a self-attention mechanism.

\subsubsection{Pyramid Fusion }

Inspired by the Feature Pyramid Network (FPN) \cite{lin2017feature}, 
we further design a Pyramid Fusion (PF) method that allows the CapsNet to represent the entities at multiple levels of granularity.
The FPN creates a pyramid of feature maps at different scales and then leverages the aggregated multi-level features
for downstream operations,
which is widely used to address the problem of scale variance in object detection and semantic segmentation tasks. 
Here, our PF builds a pyramid in the routing mechanism, capturing multiple scales and views of entities for classification. 
Taking advantage of CapsNet's ability to record the complex feature relationship, the PF is beneficial to enhance the representation ability of CapsNet.

\begin{figure}
    \centering
    \resizebox{1.0\textwidth}{!}{\includegraphics{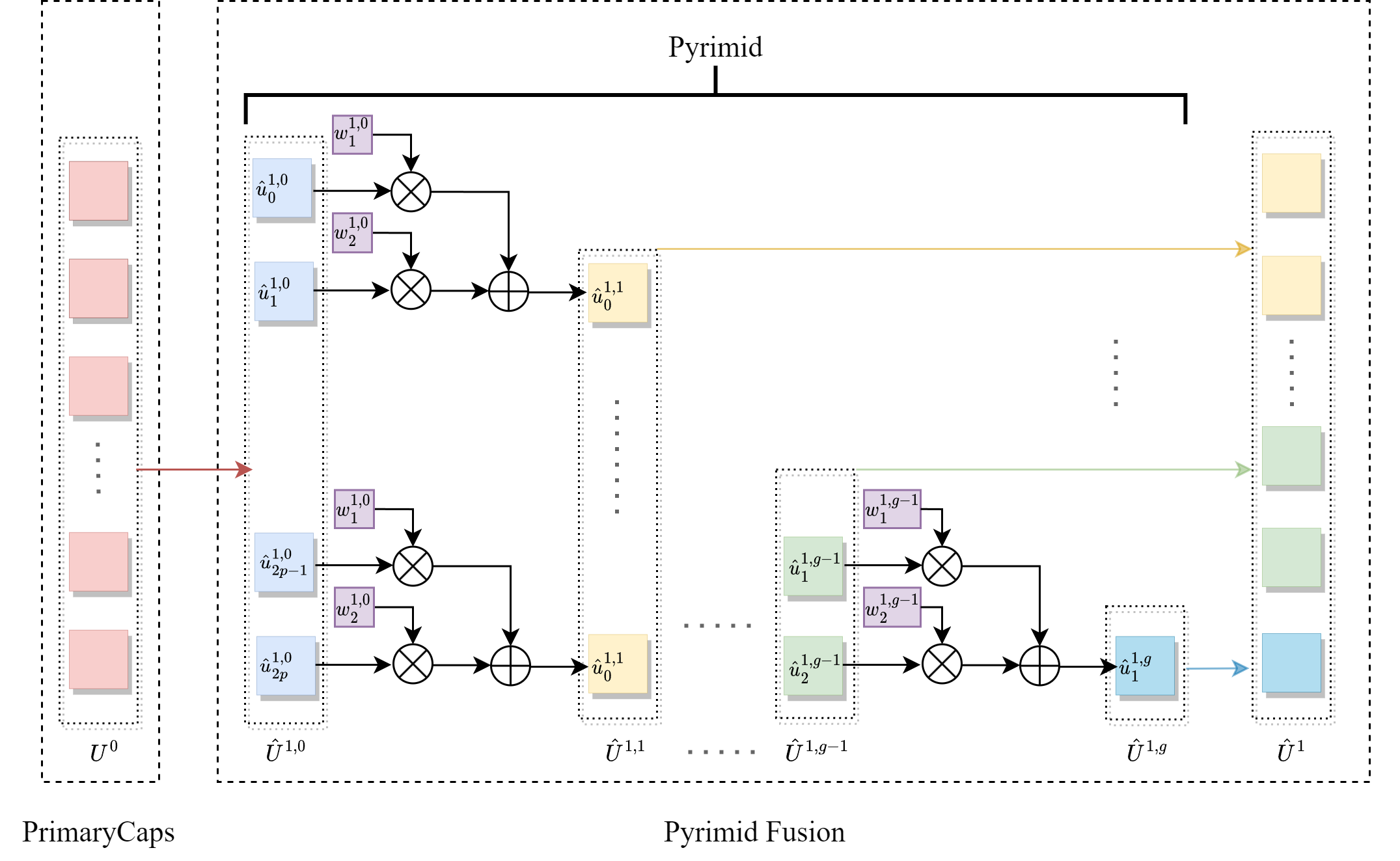}}
    \caption{The architecture of Pyramid Fusion in Multi-scale Routing method.
    In the first routing process, the output $U^{0}$ of PrimaryCap is used as the input to the pyramid. On each level of the pyramid, two adjacent prediction tensors  $\hat{u}^{1,g-1}_{2p-1}, \hat{u}^{1,g-1}_{2p}$ will be used to generate high-level ones by multiplying the sharing weights $w^{1,g-1}_{1}, w^{1,g-1}_{2}$ of the current layer. Finally, all tensors on the pyramid are concatenated together in $\hat{U}^{1}$ and passed to the next layer.
    }
    \label{cpf}
\end{figure}

The PF method builds a pyramid of prediction tensors that will be used for generating the activity tensors by using a designed merging rule.
The PF tasks activity tensors $U^{l_{c}-1}$ of previous capsule layers as input, and output the prediction tensors $\hat{U}^{l_{c}}$ of the current layer, where $l_{c} \geq 1$ represents the index of the capsule layer.
The prediction tensors on the intermediate level of the pyramid are denoted as $\hat{U}^{l_{c}, g}$, where $g$ is the index of the pyramid level and $g \geq 1$.

As shown in Figure~\ref{cpf}, taking the PF in the first routing operation as an example, 
activity tensors of PrimaryCaps, $U^0$, are copied and considered as the initial prediction tensors, $\hat{U}^{1,0}$.
The number of prediction tensors in the low-level layer (i.e., $\hat{U}^{1,0}$) and high-level layer (i.e., $\hat{U}^{1,1}$) are assumed to be $2p$ and $p$ respectively.
All the prediction tensors in the $\hat{U}^{1,0}$ are divided into $p$ groups, where two adjacent prediction tensors join into one group, such as $\hat{u}^{1,0}_{2p-1}$ and $\hat{u}^{1,0}_{2p}$.
In the merging step, the prediction tensors in $\hat{u}^{1,1}$ are generated by 
\begin{equation}
\hat{u}_{p}^{1,1}=w_{1}^{1, 0} \times \hat{u}^{1,0}_{2p-1}+ w_{2}^{1, 0} \times \hat{u}^{1,0}_{2p}\label{up-equation} ,
\end{equation}
where $w_{1}^{1, 0}$ and $w_{2}^{1, 0}$ are two shared weight matrices that are only used in the current layer.

In other words, we assign two shared weight matrices $W^{l_{c},g-1}=[w_{1}^{l_{c},g-1}, w_{2}^{l_{c},g-1}] \in \mathbb{R}^{N_{c} \times N_{c} \times 2}$ for each layer in the pyramid.
The shared weight matrices not only reduce the number of parameters but also gain global information on the pyramid for the prediction tensor generation.
As such, the merging step in the $k$-th pyramid layer is defined as 
\begin{equation}
\hat{u}_{p}^{l_{c}, g}= w_{1}^{l_{c},g-1} \times \hat{u}^{l_{c}, g-1}_{2p-1}+   w_{2}^{l_{c},g-1} \times \hat{u}^{l_{c}, g-1}_{2p}. 
\label{up-equation}
\end{equation}
After several rounds of merging operations between layers, only two prediction tensors are left on the top of the pyramid, such as $\hat{U}^{1, g}={\hat{u}_{1}^{1,g}, \hat{u}_{2}^{1,g}}$.

Finally, we concatenate all generated prediction tensors on the pyramid as the output of PF, which can be represented as:
\begin{equation}
\hat{U}^{1} = [
\hat{u}^{1,1}_{1}, \hat{u}^{1,1}_{2},\ldots \hat{u}^{1,1}_{p}, 
\ldots, 
\hat{u}_{1}^{1,g}, \hat{u}_{2}^{1,g} 
].
\label{anew}
\end{equation}
It is noteworthy that the number of activity tensors in $\hat{U}^{1} $ is the same as the initial layer $\hat{U}^{1,0}$. 
Compared with the conventional method shown in Equation~\ref{eq:prediction_vectors}, our PF operation can present and record the entities in multiple levels of granularity.

\subsubsection{Partial Connection}

\begin{figure*}
    \centering
    \includegraphics[width=01.0\linewidth]{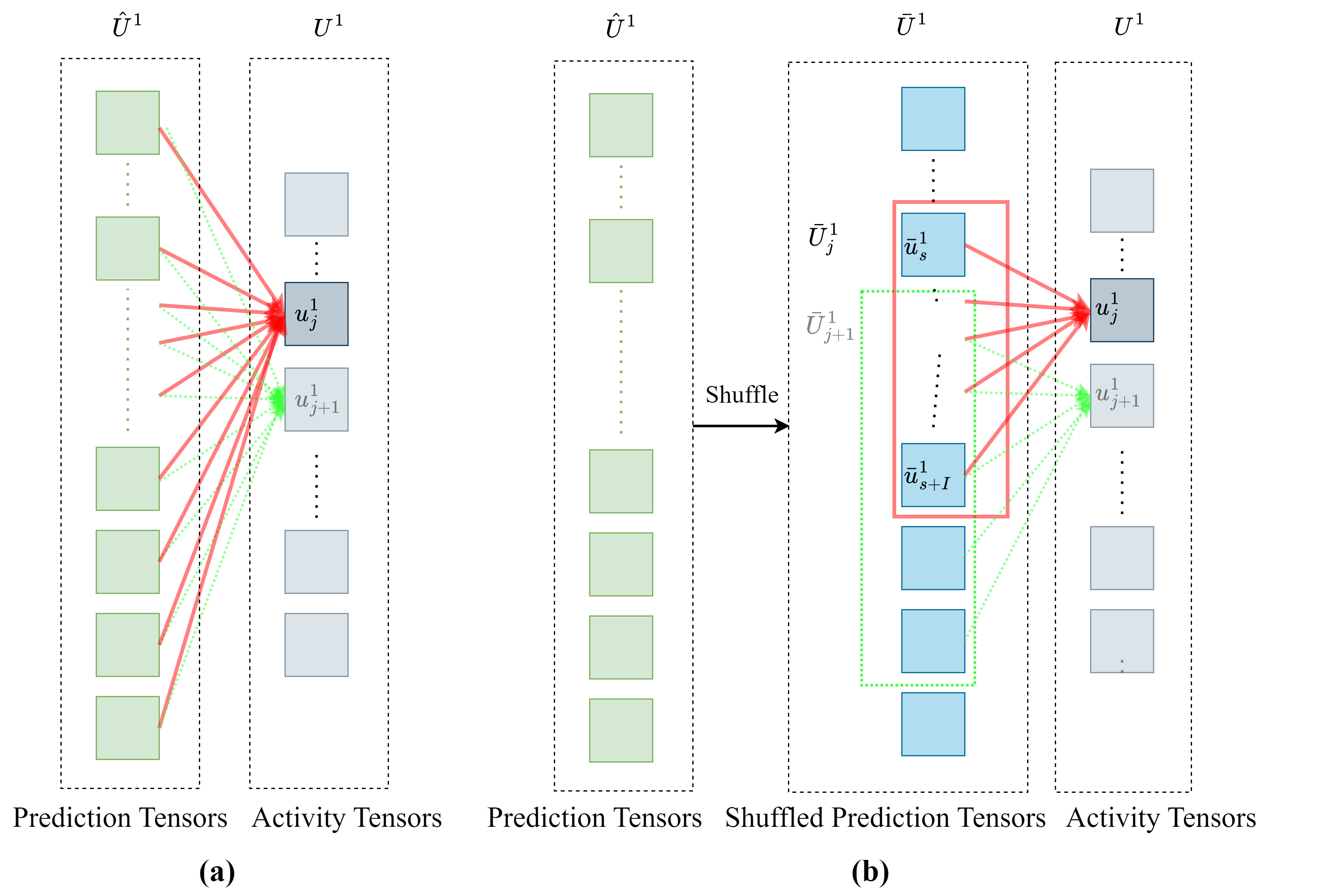}
    \caption{Two connection methods: (a) fully connection: all low-level layer capsules from $\hat{U}^1$ attend the processing of generating for every high-level capsule. (b) Partial Connection: just part of low-level layer capsules $\bar{U}_j^{1}$ attend the generation of $j$-th high-level capsule.}
    \label{partial}
\end{figure*}

In the conventional generation process of precision tensors, all capsules are organized by leveraging a fully connected method as Figure~\ref{partial} (a), where any capsules in the high-level layers need to attend to the voting processes conducted by all capsules in the low-level layers. 
As indicated in previous sections, after our PF and attentive DWT downsampling, more fine-grained spatial information and relationships are preserved in our feature map compared with previous methods. 
Therefore, we can carefully prune the full connections to reduce the computational demand. 
In this paper, we propose a partial connection method to build an efficient CapsNet, where only part of the capsules are attended to high-level capsule generation as Figure~\ref{partial} (b).

Given the prediction tensors $\hat{U}^{1} \in \mathbb{R}^{ N_{c} \times N_{c} \times 2p}$ derived from activity tensors of the low-level capsule layer, we define a sliding window method to keep the partial connection to generate activity tensors of high-level capsule layer.
Firstly, we randomly change the order of prediction tensors in $\hat{U}^{1}$, which derives a new prediction tensor set 
\begin{equation}
\bar{U}^{1}=\{\bar{u}_{1}^{1},\ldots,\bar{u}_{2p}^{1}\}.
\end{equation}
Then, a window will slide from the beginning of $\bar{U}^{1}$ to the end to keep the local and partial connection.
When the number of capsules on the high-level layer is $J$ and the goal is to generate the activity tensor of $j$-th capsule, 
the partially connected prediction tensors within the sliding window are denoted as 
\begin{equation}
\bar{U}^{1}_j=\{\bar{u}_{s}^{1},\ldots,\bar{u}_{s+I}^{1}\},
\end{equation}
where $I$ is the length of the window,
$s$ and $s+I$ represent the star and end index of the window respectively,
$s= a \times j$ and $s\leq 2p-N$,
and $a = \lfloor \frac{2p}{J} \rfloor$ is the stride of sliding window.

We employ the self-attention mechanism \cite{DBLP:journals/corr/VaswaniSPUJGKP17} to calculate the coupling coefficients for generating activity tensors of high-level capsules.
Given a window of prediction tensors $\bar{U}^{1}_j$, attention coefficients $C_{j \mid i}$ produced by the self-attention mechanism determine which low-level capsules should pay more attention. The higher scores correspond to higher weights so that the informative elements for classification receive higher attention.
Therefore, the $C_{j \mid i}$ is reasonable to act as coupling coefficients in the routing algorithm.

The $\bar{U}^{1}_j \in \mathbb{R}^{N_{c} \times N_{c} \times I}$ is reshaped to a tensor with dimension of $I \times N_{c}^{2} \times 1$. Here, we abuse the notation $\bar{U}^{1}_j$ to denote the obtained tensor.
The query and key tensors of self-attention are obtained by applying linear transformations which are shown as 
\begin{equation}
    Q_j = \bar{U}^{1}_j \times W_{Q_j}, \quad K_j = \bar{U}^{1}_j \times W_{K_j} ,
\end{equation}
where $Q_i, K_i \in \mathbb{R}^{ I \times N_{c}^{2} \times 1}$, and $W_{Q_j}, W_{K_j} \in \mathbb{R}^{N_{c}^{2} \times N_{c}^{2} \times 1}$ are learnable weight matrices.
Subsequently, we gain the attention coefficients $C_{j|i} \in R^{I\times I \times 1}$ by following equation:
\begin{equation}
C_{j|i}=\text{Attention}(Q_i,K_i) =\text{softmax}\left(\frac{Q_iK_i^T}{\sqrt{N_c}}\right).
\label{attention-eq}
\end{equation}
The $C_{j|i}$ 
measures the importance of prediction tensors in $\bar{U}^{1}_j$ to assign the weights.
Accordingly, the voting vectors for $j$-th capsule in the high-level layer are defined as:
\begin{equation}
V_{j|i}= C_{j|i} \times \bar{U}^{1}_j.
\end{equation}
%
We reshape $V_{j|i} \in \mathbb{R}^{I \times N_{c}^{2} \times 1}$ into $V_{j|i} \in \mathbb{R}^{N_c \times N_c \times I}$, and  $V_{j|i}=\{v_1,v_2,...,v_{I}\}$. There are $I$ capsules in $V_{j|i}$, and the capsules belong to $V_{j|i}$ has encoded the relationship among entities and can generate the high layer capsules as:
\begin{equation}
u_{j} = \frac{1}{I} \sum_{i=1}^{I} v_{i}.
\label{routing-eq}
\end{equation}
Finally, we get the activity tensor set $U^{1}$ that is shown as
\begin{equation}
U^{1} = \{u^{1}_{1},u^{1}_{2},\ldots,u^{1}_{J}\}.
\label{final}
\end{equation}

In terms of learning objective, Notably, we only leverage the loss term $L_{\text {margin}}$ in Equation~\ref{op:loss} for optimization here.
In addition, squashing function in Equation~\ref{squash} is also not used, since it will lead to the whole model being hard to converge quickly in practice.

\section{Results}

\subsection{Experimental Setting}

All experiments were completed on CUDA 17.1, Python 3.8, and Pytorch 1.8.1. At the same time, all GPUs used are single RTX 3090.
We utilize the $ResNet$-18 as the backbone for feature extraction.
After each attentive DWT downsampling operation, the shape of feature maps is changed from $N \times N \times K$ to $\frac{N}{2} \times \frac{N}{2} \times 4K$.
Our CapsNet includes three capsule layers, where the number of capsules in each layer is $\{64,64,n\}$ respectively, and $n$ equals to the number of classes of the dataset.  
For the activity tensor set $U^{l_{c}} \in \mathbb{R}^{N_c \times N_c \times 2p}$ of each capsule layer, 
the $2p$ represents the number of capsules, the shape for each activity tensor $u^{l_{c}}$ is $N_c \times N_c$, and the $N_{c}$ is uniformly set as $4$ for all methods on different datasets.
Additionally, the number of capsules in $V_{j|i}$ is $9$ in the multi-scaling routing method.

To enhance training efficiency and mitigate the risk of overfitting due to limited training samples, we incorporate early stopping and a dynamic learning rate strategy for DWT-CapsNet. Specifically, the training process will stop if the validation loss no longer decreases for 50 consecutive epochs. Additionally, the learning rate is reduced by half if the validation loss does not decrease for 10 consecutive epochs. We set the maximum number of training epochs to 270, with an initial learning rate of 0.001.

During the testing phase, we employ several evaluation metrics to quantitatively assess the classification performance of the proposed models. These metrics include Overall Accuracy (OA), Average Accuracy (AA), and the Kappa Coefficient (KC). These metrics provide a comprehensive evaluation of the models' performance in HSI classification. To evaluate the model efficiency, we introduce training time, testing time, the number of parameters, and Floating-point operations per second (FLOPs).

To illustrate the effectiveness of our method, we compare DWT-CapsNet with several baselines: 
(1) SVM \cite{melgani2004classification} and (2) 3D-CNN \cite{srivastava2015training} employ the vanilla Support Vector Machine and 3D convolutional network respectively for classification. 
(3) Spectral-Spatial Residual Network (SSRN) \cite{zhang2019three}: the 3D densely connected convolutional network is used for extracting spectral–spatial information for HSI classification. (4) Deep Feature Dense Network (DFDN) \cite{li2019deep}: DFDN fuses the low, middle, and high-level features. 
(5): Nonlocal CapsNet (NLCapsNet) \cite{lei2021non}: Non-local attention mechanism is embedded into capsules to focus on important spatial information. (6) DC-CapsNet \cite{lei2021hyperspectral}: 3D convolutional capsule layers are leveraged for building a deep convolutional CapsNet. (7) MS-CapsNet \cite{lei2022multiscale}: Multiscale feature aggregation mechanism helps CapsNet to gain both local and global spatial features and integrated multiscale features, which improves model robustness.


\subsection{Datasets}


To assess the effectiveness of these models, we employ four following HSI datasets for evaluation: 

\textbf{Kennedy Space Center (KSC) Dataset (Florida, 1996):} The KSC dataset was acquired in Florida in 1996 and comprises images with dimensions of 512x614 pixels, consisting of 176 bands. This dataset encompasses 13 distinct categories of ground objects and its input size is $11\times 11$.

\textbf{Pavia University (UP) Dataset (Pavia, Italy):} UP Dataset was collected in northern Italy, the UP dataset collects images with dimensions of 610x340 pixels and includes 103 bands. It is annotated with ground-truth information for 9 different classes, and the input size is $9\times 9$.

\textbf{Salinas Dataset (Salinas Valley, California):} The SA dataset was gathered over the Salinas Valley in California, containing images sized at 512x217 pixels and composed of 204 bands. It encompasses 16 distinct land-cover classes and its patch size is $9 \times 9$.

\textbf{WHU-Hi-LongKou (LK) Dataset (Longkou Town, China, 2018):} This dataset was acquired in Longkou Town, Hubei province, China in 2018, the LK dataset comprises images with dimensions of 550x400 pixels. It includes 270 effective bands and covers 9 land-cover classes with patch size $9\times 9$.

Following a standard evaluation protocol \cite{lei2022multiscale},
for all experiments in this section, a fraction of each dataset is allocated for training and validation, while the remaining labeled samples are reserved for testing the performance of the DWT-CapsNet model. Specifically, $5\%$ of the KSC dataset, $1\%$ of the UP and SA datasets, and $0.2\%$ of the LK dataset are designated for the training and validation sets.

\subsection{Main Results}

Table \ref{total} has shown that CapsNet can gain better AA, OA, and K compared with other traditional methods such as SVM and CNN methods on four different datasets. Meanwhile, the DWT-CapsNet achieves state-of-the-art in all kinds of capsules on AA, OA, and KC.

On the KSC datasets, DWT-CapsNet improves almost 18\%, 12\%, 6\% compared with SVM, 3D-CNN, and SSRN on OA metrics. Compared with other capsule networks such as NLCapsNet, DC-capsNet, and MS-CpasNet, DWT-CapsNet is greater than NLCapsNet and DC-CapsNet by about 7\% and 5\%.  For the MS-Capsule, our method has slight advantages (about 1.2\%). For other metrics (AA, KC), DWT-CapsNet precedes other models from 1.5\% to 20\% inconsistently. The performance comparison on the LK dataset also reveals DWT-CapsNet's significant superiority over other CapsNet. Specifically, DWT-CapsNet excels in terms of OA, achieving an impressive result (97.57\% on OA) on the complex LK dataset. In contrast, NLCapsNet lags behind with an OA of 95.41\%. Additionally, DC-CapsNet falls short at 95.63\%.

\begin{table*}[]
\caption{Classification results from different models with the Kennedy Space Center, Pavia University,
Salinas, and WHU-Hi-LongKou datasets.}
\centering
\tiny
\begin{tabular}{c@{\ \ }c@{\ \ }c@{\ \ }c@{\ \ }c@{\ \ }c@{\ \ }c@{\ \ }c@{\ \ }c@{\ \ }c}
\toprule
Dataset   & Models & SVM          & 3D-CNN       & SSRN         & DFDN         & NLCapsNet    & DC-CapsNet   & MS-CapsNet   & DWT-CapsNet \\ 
\midrule
\multirow{3}{*}{KSC} & OA(\%) & 81.83$\pm$0.04 & 87.65$\pm$1.89 & 93.25$\pm$1.25 & 88.43$\pm$0.88 & 93.21$\pm$0.79 & 95.97$\pm$1.16 & 96.67$\pm$0.63 & \textbf{99.77$\pm$0.42} \\ 
                     & AA(\%) & 78.36$\pm$2.33 & 85.69$\pm$2.40 & 91.62$\pm$1.02 & 87.58$\pm$1.44 & 92.00$\pm$0.95 & 93.43$\pm$1.84 & 96.60$\pm$9.71 & \textbf{99.59$\pm$1.62} \\ 
                     & KC$\times$100  & 79.73$\pm$0.05 & 86.24$\pm$2.11 & 92.51$\pm$1.41 & 87.11$\pm$0.96 & 92.43$\pm$0.88 & 95.51$\pm$1.29 & 97.41$\pm$0.69 & \textbf{99.61$\pm$0.59}\\ 
\midrule
\multirow{3}{*}{UP}  & OA(\%) & 78.53$\pm$0.74 & 86.55$\pm$0.97 & 95.23$\pm$0.57 & 88.77$\pm$1.47 & 89.79$\pm$1.96 & 96.71$\pm$0.38 & 98.41$\pm$0.58 & \textbf{99.19$\pm$1.03} \\ 
                     & AA(\%) & 69.94$\pm$0.91 & 82.76$\pm$2.07 & 93.74$\pm$0.52 & 86.06$\pm$1.36 & 87.87$\pm$2.05 & 95.51$\pm$0.41 & 97.59$\pm$0.81 & \textbf{99.25$\pm$1.49} \\ 
                     & KC$\times$100  & 70.68$\pm$0.92 & 81.96$\pm$1.29 & 93.75$\pm$0.77 & 84.96$\pm$2.01 & 86.42$\pm$1.64 & 95.63$\pm$0.50 & 97.89$\pm$0.77 & \textbf{99.57$\pm$1.23} \\ 
\midrule
\multirow{3}{*}{SA}  & OA(\%) & 83.69$\pm$1.39 & 87.81$\pm$1.72 & 95.29$\pm$0.26 & 88.80$\pm$1.78 & 93.17$\pm$1.61 & 97.14$\pm$0.32 & 98.20$\pm$0.01 & \textbf{98.39$\pm$0.71} \\ 
                     & AA(\%) & 86.34$\pm$2.05 & 92.18$\pm$1.81 & 97.40$\pm$0.13 & 90.53$\pm$2.24 & 94.69$\pm$0.71 & 98.06$\pm$0.43 & 98.60$\pm$0.76 & \textbf{99.01$\pm$0.33} \\ 
                     & KC$\times$100  & 81.75$\pm$1.57 & 86.36$\pm$1.96 & 94.76$\pm$0.28 & 87.51$\pm$1.99 & 92.39$\pm$1.79 & 96.82$\pm$0.35 & 98.00$\pm$0.01 & \textbf{98.70$\pm$0.76} \\ 
\midrule
\multirow{3}{*}{LK}  & OA(\%) & 82.89$\pm$0.35 & 92.60$\pm$0.88 & 95.54$\pm$0.48 & 92.54$\pm$0.55 & 92.16$\pm$0.91 & 94.83$\pm$0.47 & 97.35$\pm$0.45 & \textbf{97.57$\pm$1.01} \\ 
                     & AA(\%) & 45.62$\pm$0.52 & 83.13$\pm$1.96 & 93.75$\pm$0.22 & 80.99$\pm$2.17 & 78.88$\pm$1.58 & 86.56$\pm$0.44 & 93.03$\pm$1.52 & \textbf{95.41$\pm$1.14} \\ 
                     & KC$\times$100  & 76.45$\pm$0.48 & 92.23$\pm$1.12 & 94.10$\pm$0.64 & 90.15$\pm$0.74 & 89.61$\pm$1.20 & 93.19$\pm$0.63 & 96.52$\pm$0.60 & \textbf{96.52$\pm$0.79} \\ 
\bottomrule
\end{tabular}
\label{total}

\end{table*}

On the SA dataset, CapsNet, particularly DC-CapsNet, and MS-CapsNet, demonstrate remarkable performances, achieving high OA of approximately 97.14\% and 98.20\%, respectively. These results suggest capsule networks capture intricate patterns and relationships within the SA dataset, outperforming the traditional SVM model that has a weak OA of approximately 83.69\%. By achieving an impressive OA of 99.77\%, DWT-CapsNet stands out as excellent work for HSI tasks. This outstanding accuracy shows the model's ability to effectively capture intricate spatial relationships and complex features within the SA dataset, surpassing other capsule methods. On the University of Pavia (UP) dataset, DWT-CapsNet reaffirms its superiority.  DWT-CapsNet attains an impressive OA of 99.19\%, while DC-CapsNet slightly lags with an OA of 98.41\%. This gap underscores DWT-CapsNet's consistent and robust performance in accurately categorizing UP dataset data, which is vital for land-use classification and urban planning applications. Additionally, in comparison to NLCapsNet on the UP dataset, DWT-CapsNet reaches an OA of 99.19\%, while NLCapsNet achieves only 89.79\%. This significant performance difference highlights DWT-CapsNet's power in handling the intricacies of the UP dataset, making it an ideal choice for tasks involving urban and agricultural land classification.

\subsection{Ablation Studies}
\begin{table}[]
\caption{Ablation experiment for DWT-CapsNet. Our attentive DWT downsampling and multi-scale routing are denoted as A-DWT-D and M-SR respectively.}
\centering
\small
\begin{tabular}{cccccc}
\toprule
Dataset              & Models & Our CNN & CNN + A-DWT-D & CNN + M-SR & DWT-CapsNet \\ \midrule
\multirow{3}{*}{KSC} & OA(\%) & 90.49   & 91.19   & 99.66       & 99.77      \\
                     & AA(\%) & 93.91   & 92.77   & 99.53       & 99.59      \\
                     & KC$\times$ 100  & 94.51   & 94.58   & 97.91       & 99.61      \\ \midrule
\multirow{3}{*}{UP}  & OA(\%) & 93.79   & 93.34   & 99.19       & 99.15      \\
                     & AA(\%) & 94.15   & 95.71   & 98.51       & 99.29      \\
                     & KC$\times$ 100  & 92.59   & 93.49   & 99.17       & 99.57      \\ \midrule
\multirow{3}{*}{SA}  & OA(\%) & 93.94   & 95.01   & 97.98       & 98.39      \\
                     & AA(\%) & 95.03   & 96.03   & 98.95       & 99.01      \\
                     & KC$\times$ 100  & 93.97   & 93.91   & 97.41       & 98.7       \\ \midrule
\multirow{3}{*}{LK}  & OA(\%) & 92.93   & 93.39   & 97.51       & 97.57      \\
                     & AA(\%) & 91.65   & 92.71   & 95.69       & 95.41      \\
                     & KC$\times$ 100  & 90.97   & 94.14   & 96.34       & 96.52      \\ 
\bottomrule
\end{tabular}
\label{table2}
\end{table}
In this comprehensive analysis, we will explore how each component contributes to the accuracy of classification models on four distinct datasets: KSC, UP, SA, and LK. We aim to elucidate how the integration of various components - namely, CNN backbone, attentive DWT downsampling, Muilt-scaling routing, and the completed DWT-CapsNet model - straightforwardly influences the models' accuracy from Table \ref{table2}.

Our analysis begins with the performance of the baseline CNN model. This foundational component provides a decent starting point for classification across all datasets. On the KSC dataset, the CNN backbone achieves an OA of 90.49\%, with an AA of 93.91\%. UP dataset exhibits similar consistency, with an OA of 93.79\% and an AA of 94.15\%. SA dataset follows suit, displaying an OA of 93.94\% and an AA of 95.03\%. Even the challenging LK dataset maintains respectable accuracy, with an OA of 92.93\% and an AA of 91.65\%. This steady performance underscores the reliability of CNN as a foundational element.

We observe notable improvements as we introduce attentive DWT downsampling, especially on the SA dataset. Adding DWT leads to an increase in OA from 93.94\%  to 95.01\%, and the AA exhibits a minor increase from 95.03\% to 96.03\%. This trend highlights attentive DWT downsampling's positive influence on the model's ability to capture useful features, which is particularly beneficial for complex datasets. This component demonstrates its role in enhancing the model's capacity to reduce object information loss. Including multi-scale routing behind the CNN model leads to a transformative leap in accuracy. On the KSC dataset, the combined CNN+muilt-scale routing model propels OA to an impressive 99.66\%, while AA reaches 99.53\%. This remarkable enhancement demonstrates multi-scale routing's unique ability to record the relationship of objects. The UP dataset also shows substantial improvements, with OA surging from 93.34\%  to 99.19\%, highlighting multi-scale routing's consistency in enhancing accuracy across datasets. Finally, we reach the height of our analysis with DWT-CapsNet, where the integration of CNN, DWT, and CapsNet combines their strengths to create a powerful, integrated model. On the KSC dataset, DWT-CapsNet achieves an exciting OA of 99.77\%, showing a substantial leap from the baseline CNN model's 90.49\%. The AA also reaches an impressive 99.59\%. This integrated model's capacity to capture intricate patterns and multi-resolution features is demonstrated.  These results underscore the cumulative effect of the components within DWT-CapsNet, demonstrating that the whole is greater than the sum of its parts.

\begin{figure}[ht]
	\centering
	\subfloat[]{\includegraphics[width=.15\linewidth]{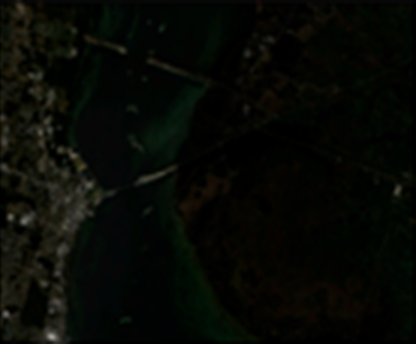}}\hfill 
        \subfloat[]{\includegraphics[width=.15\linewidth]{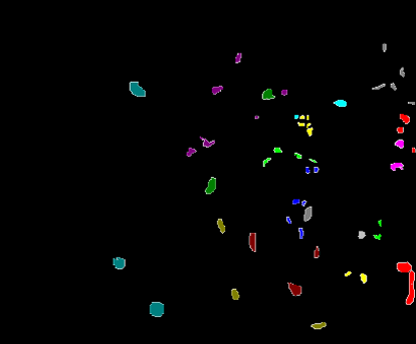}}\hfill 
	\subfloat[]{\includegraphics[width=.15\linewidth]{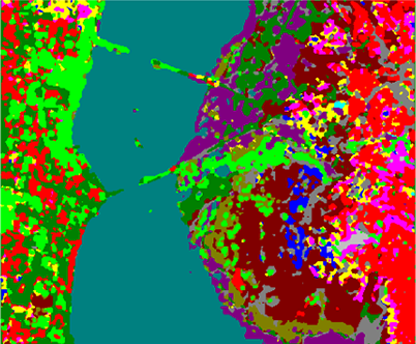}}\hfill
	\subfloat[]{\includegraphics[width=.15\linewidth]{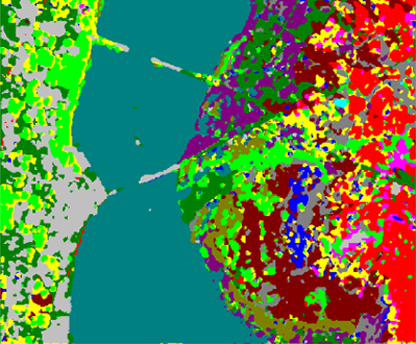}}\hfill 
        \subfloat[]{\includegraphics[width=.15\linewidth]{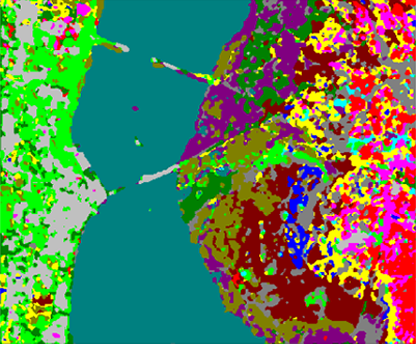}}\hfill 
	\subfloat[]{\includegraphics[width=.15\linewidth]{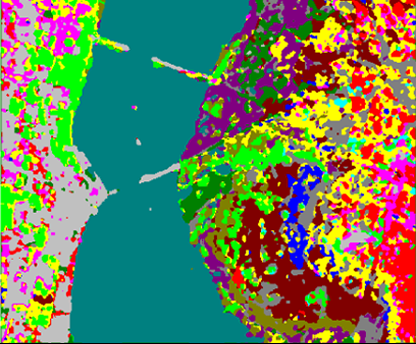}}\hfill
    \newline
        \subfloat{\includegraphics[width=0.6\linewidth]{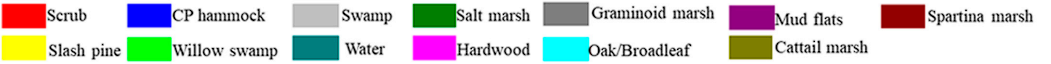}}\
	\caption{Different models implemented on the  Kennedy Space Center dataset: (a) false-color image, (b) ground-truth label, (c) classification result with backbone, (d) classification result with backbone + attentive DWT downsampling, (e) classification result with backbone + multi-scale routing, (e) classification result with backbone + Capsule, (f) classification result with DWT-CapsNet.}
    \label{KSC}

\end{figure}

\begin{figure}[htbp]
	\centering
	\subfloat[]{\includegraphics[width=.13\linewidth]{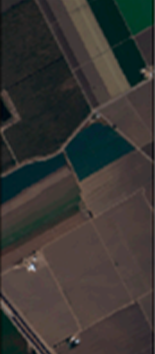}}\hfill 
    \subfloat[]{\includegraphics[width=.13\linewidth]{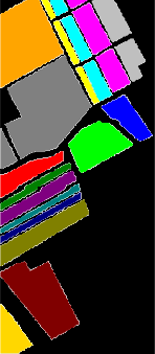}}\hfill 
	\subfloat[]{\includegraphics[width=.13\linewidth]{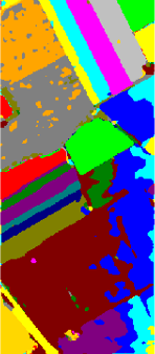}}\hfill
	\subfloat[]{\includegraphics[width=.13\linewidth]{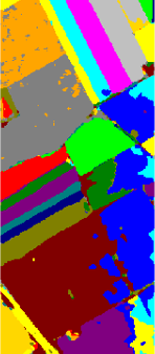}}\hfill 
    \subfloat[]{\includegraphics[width=.13\linewidth]{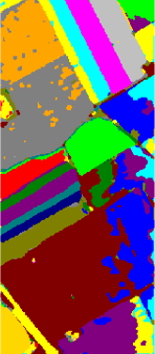}}\hfill 
	\subfloat[]{\includegraphics[width=.13\linewidth]{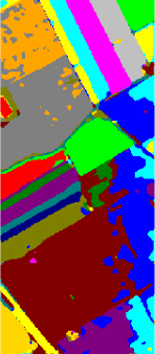}}\hfill
	\subfloat{\includegraphics[width=.13\linewidth]{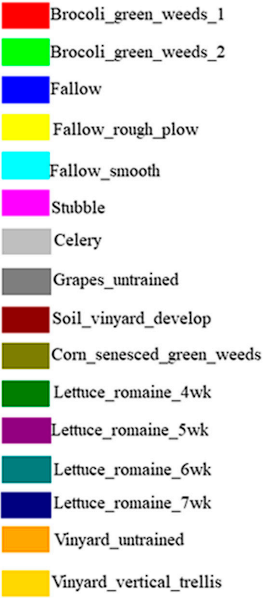}}\hfill 
	\caption{Different models implemented on the Salinas dataset:(a) false-color image, (b) ground-truth label, (c) classification result with backbone, (d) classification result with backbone + attentive DWT downsampling, (e) classification result with backbone + multi-scale routing, (e) classification result with backbone + Capsule, (f) classification result with DWT-CapsNet.}
    \label{SA}

\end{figure}

\subsection{Visualization for Predictions}

\begin{figure}[htbp]
	\centering
	\subfloat[]{\includegraphics[width=.13\linewidth]{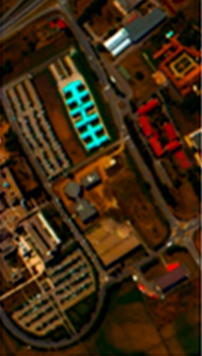}}\hfill 
    \subfloat[]{\includegraphics[width=.13\linewidth]{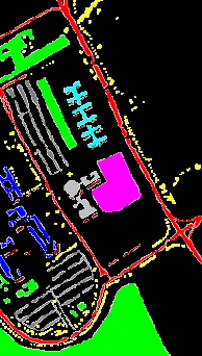}}\hfill 
	\subfloat[]{\includegraphics[width=.13\linewidth]{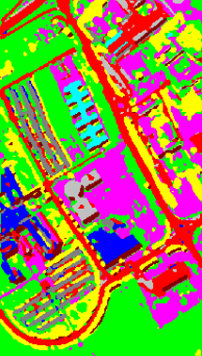}}\hfill
	\subfloat[]{\includegraphics[width=.13\linewidth]{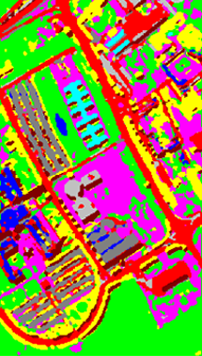}}\hfill 
    \subfloat[]{\includegraphics[width=.13\linewidth]{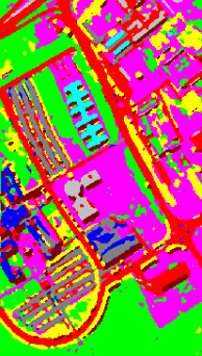}}\hfill 
	\subfloat[]{\includegraphics[width=.13\linewidth]{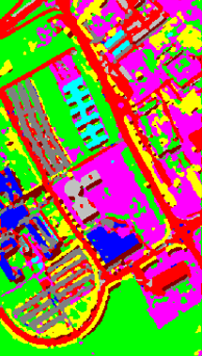}}\hfill
 	\subfloat{\includegraphics[width=.08\linewidth]{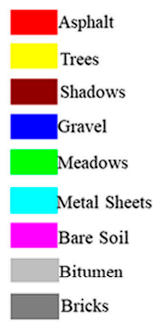}}\hfill
	\caption{Different models implemented on the Pavia University dataset: (a) false-color image, (b) ground-truth label, (c) classification result with backbone, (d) classification result with backbone + attentive DWT downsampling, (e) classification result with backbone + multi-scale routing, (e) classification result with backbone + Capsule, (f) classification result with DWT-CapsNet.}
    \label{UP}

\end{figure}

\begin{figure}[htbp]
	\centering
	\subfloat[]{\includegraphics[width=.13\linewidth]{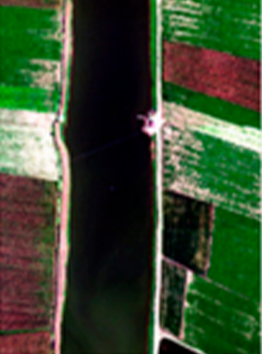}}\hfill 
    \subfloat[]{\includegraphics[width=.13\linewidth]{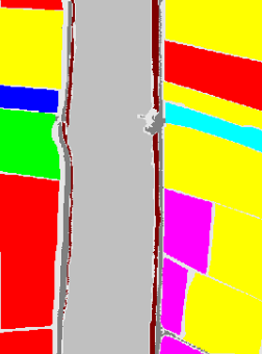}}\hfill 
	\subfloat[]{\includegraphics[width=.13\linewidth]{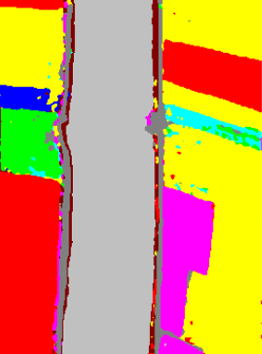}}\hfill
	\subfloat[]{\includegraphics[width=.13\linewidth]{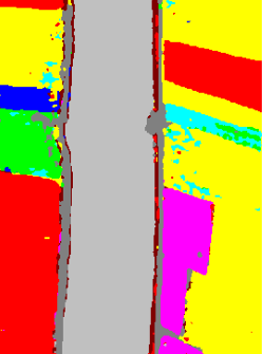}}\hfill 
    \subfloat[]{\includegraphics[width=.13\linewidth]{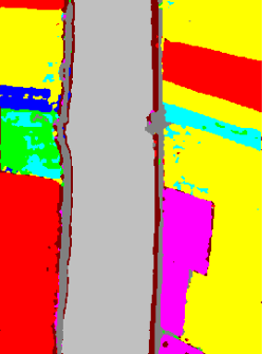}}\hfill 
	\subfloat[]{\includegraphics[width=.13\linewidth]{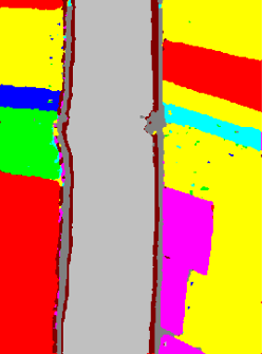}}\hfill
 	\subfloat{\includegraphics[width=.11\linewidth]{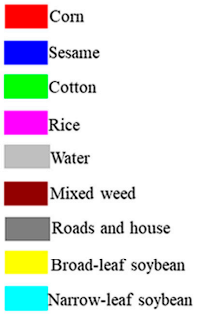}}\hfill
	\caption{Different models implemented on the WHU-Hi-LongKou dataset: (a) false-color image, (b) ground-truth label, (c) classification result with backbone, (d) classification result with backbone + attentive DWT downsampling, (e) classification result with backbone + multi-scale routing, (e) classification result with backbone + Capsule, (f) classification result with DWT-CapsNet.}
    \label{LK}

\end{figure}

Based on classification results, we make the visualization of classification results as Figure~\ref{KSC}-\ref{LK}. The whole DWT-CapsNet can make the images smoother, and each building structure is clear even though there is a small object in the image. The DWT-CapsNets' images are most similar to the ground-truth label. Therefore, it also gains better classification results. For the ablation parts, we can find that the attentive DWT downsampling can improve spatial information for object shapes. Especially for Figure~\ref{SA}, the backbone + attentive DWT downsamplings show smoother than the backbone only, which matches the ground-truth label. The attentive DWT downsampling helps the model also reduce the wrong area on the UP dataset to improve the classification results. In addition,  our multi-scale routing is also effective, as Figure~\ref{UP} shows. The blue building is more apparent in Figure~\ref{UP} (e) compared with (a). Therefore, it is evident that the DWT and multi-scale routing can improve the classification results and prediction image quality.

\subsection{Efficiency of Methods}

Compared with other methods, DWT-CapsNet is efficient and occupies less computing resources.  One striking highlight is DWT-CapsNet's remarkable efficiency for testing time from Table~\ref{test_time}. Across all datasets, it consistently outperforms other models regarding testing time. For example, DWT-CapsNet takes only 11.52 seconds on the LK dataset for testing, while its closest competitor takes 121.66 seconds. This significant gap in testing efficiency is particularly crucial for real-time applications, where quick decision-making is vital.  Besides, DWT-CapsNet stands out for its efficient use of parameters. It balances model complexity and performance, maintaining a relatively low number of parameters compared to some models. For instance, DWT-CapsNet has 506.655K parameters on the UP dataset, significantly fewer than its counterparts. This efficiency is advantageous for deployment in resource-constrained environments. This comprehensive performance makes it suitable for a wide range of applications, from satellite image analysis (KSC and UP) to landing usage classification (LK) and urban planning (SA). DWT-CapsNet's ability to adapt to different data characteristics makes it a universal and dependable choice for data-driven tasks across various domains.

\begin{table*}[]
\caption{Training, test time, and parameters under different models.}
\centering
\tiny
\begin{tabular}{c@{\ \ }c@{\ \ }c@{\ \ }c@{\ \ }c@{\ \ }c@{\ \ }c@{\ \ }c@{\ \ }c@{\ \ }c}
\toprule
\text{Dataset} & \text{Methods} & \text{3D-CNN} & \text{SSRN} & \text{DFDN} & \text{NLCapsNet} & \text{DC-CapsNet} & \text{MS-CapsNet} & \text{MS-CapsNet-WI} & \text{DWT-CapsNet} \\ \midrule
\multirow{3}{*}{KSC} & \text{Train (s)} & 36.41 & 60.08 & 798.56 & 1492.91 & 98.40 & 143.97 & 269.66 & 296.4 \\ 
 & \text{Test (s)} & 1.54 & 5.29 & 35.49 & 55.50 & 2.96 & 10.01 & 10.09 & 11.52 \\ 
 & \text{Parameters} & 2,087,553 & 309,845 & 1,244,410 & 6,068,096 & 409,728 & 716,864 & 716,864 & 424.665 \\ \midrule
\multirow{3}{*}{UP} & \text{Train (s)} & 46.24 & 67.65 & 659.64 & 1441.15 & 47.42 & 132.58 & 307.78 & 298.6 \\ 
 & \text{Test (s)} & 8.67 & 11.39 & 151.76 & 323.62 & 20.01 & 61.77 & 66.23 & 57.39 \\ 
 & \text{Parameters} & 832,349 & 199,153 & 1,239,922 & 4,429,696 & 309,248 & 654,272 & 654,272 &506.655 \\ \midrule
\multirow{3}{*}{SA} & \text{Train (s)} & 61.44 & 125.94 & 1562.30 & 3040.28 & 129.61 & 283.18 & 420.32 & 397.67 \\ 
 & \text{Test (s)} & 18.86 & 26.63 & 373.50 & 778.01 & 32.65 & 87.96 & 92.87 & 124.99 \\ 
 & \text{Parameters} & 2,401,756 & 352,928 & 1,247,776 & 7,296,896 & 454,272 & 760,384 & 760,384 & 577,465 \\ \midrule
\multirow{3}{*}{LK} & \text{Train (s)} & 35.29 & 80.62 & 2632.34 & 730.90 & 57.67 & 125.84 & 129.51 & 130.42 \\ 
 & \text{Test (s)} & 60.41 & 121.66 & 6011.48 & 1633.44 & 131.30 & 378.15 & 329.45 & 332.13 \\ 
 & \text{Parameters} & 3,497,949 & 454,129 & 1,239,922 & 4,429,696 & 501,632 & 675,648 & 675,648 & 443.467 \\ 
\bottomrule
\end{tabular}%
\label{test_time}
\end{table*}

Table~\ref{flops} compares four distinct CapsNet models based on their computational complexity measured in FLOPs. The DWT-CapsNet model is the most computationally efficient among the listed models, necessitating only 88.6 million FLOPs. This makes it an attractive choice for scenarios with limited computational resources.
The Dynamic is the original capsule routing \cite{sabour2017dynamic}, we utilize the same structure with DWT-CapsNet to keep a fair comparison. Firstly, the Dynamic model exhibits relatively lower computational demands, with an estimated 179.6 million FLOPs. This suggests it is a computationally lightweight option for tasks where resource efficiency is paramount.
In contrast, the DC-CapsNet model significantly surpasses the Dynamic model in computational complexity, boasting approximately 399.6 million FLOPs. This indicates a considerably higher demand for computational resources during inference.
The MI-CapsNet model falls between these extremes, with around 297.4 million FLOPs. It is more computationally intensive than Dynamic but less so than DC-CapsNet.

In summary, DWT-CapsNet's superiority in testing efficiency, competitive training times, resource-efficient parameter usage, FLOPs, and robust performance across diverse datasets make it a standout model in this comparison. Its strengths in both speed and accuracy underscore its potential for real-world applications where efficiency and reliability are paramount. As data plays a crucial role in various industries, models like DWT-CapsNet that can deliver high-quality results efficiently will likely see increased adoption in practical, resource-constrained scenarios.

\begin{table}[]
\caption{Flops of different models on UP dataset.}
\centering
\small
\begin{tabular}{ccccc}
\toprule
Model & Dynamic & DC-CapsNet & MI-CapsNet & DWT-CapsNet \\ \midrule
FLOPs(M) & 179.6   & 399.6    & 297.4      & 88.6       \\ 
\bottomrule
\end{tabular}
\label{flops}
\end{table}


\subsection{Sensitivity Analysis}

\begin{figure}[htbp]
\centering
\subfloat[]{\includegraphics[width=.5\linewidth]{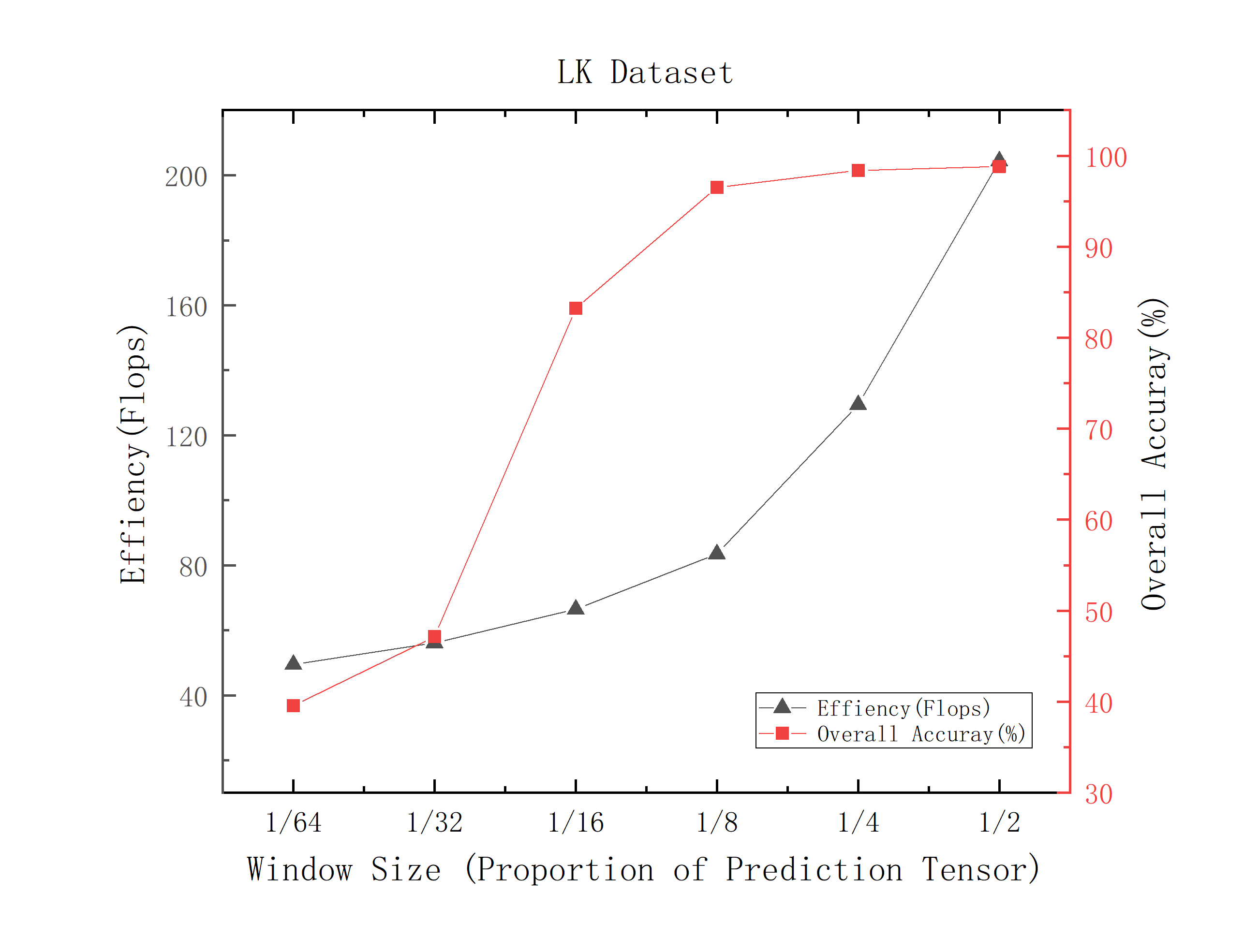}}\hfill 
\subfloat[]{\includegraphics[width=.5\linewidth]{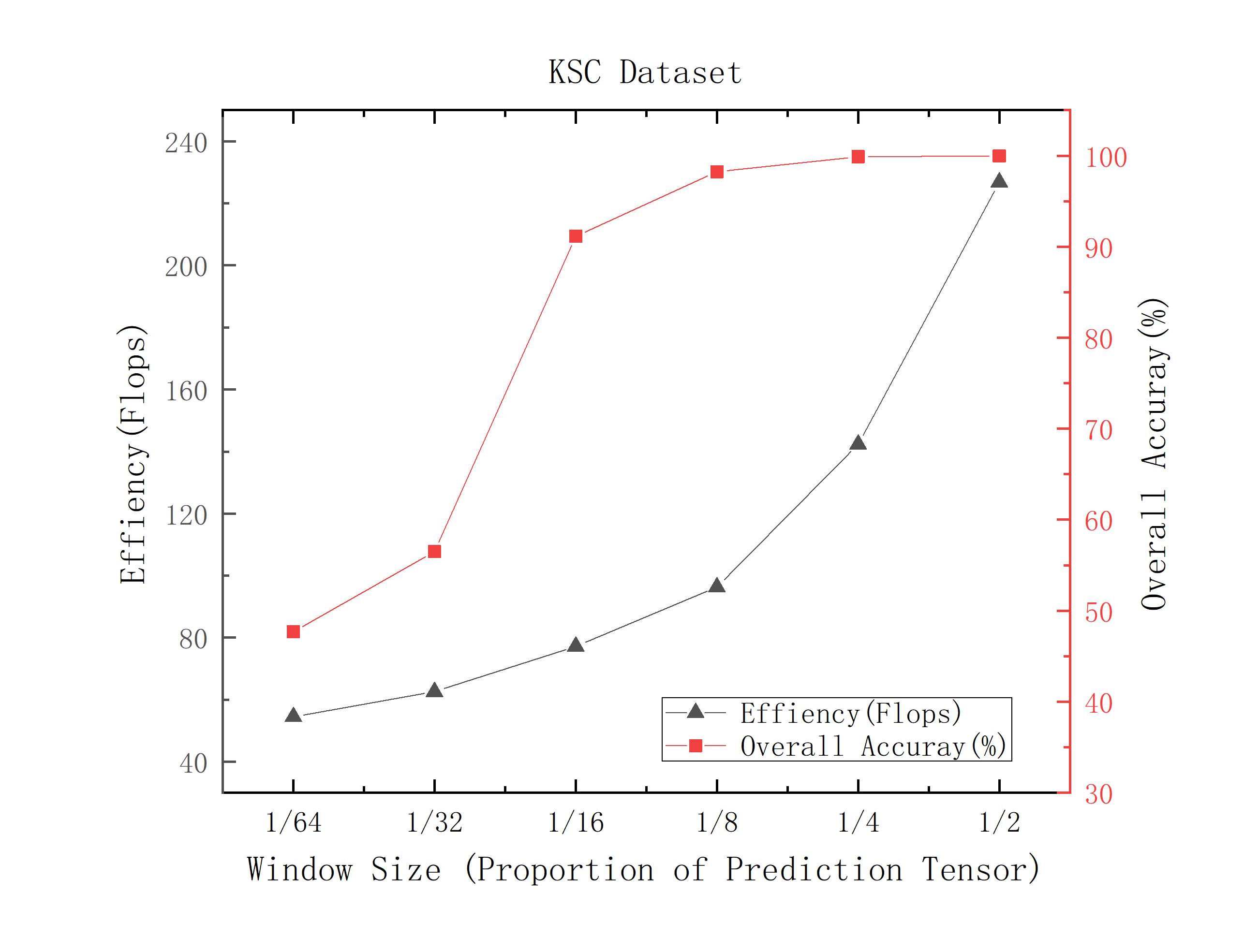}}\hfill 
	\caption{Sensitivity Analysis for window size of our partial connection mechanism on (a) LK dataset and (b) KSC dataset.}
    \label{Sensitivity}
\end{figure}

In Figure \ref{Sensitivity}, we conducted a sensitive analysis for the window size of our partial connection mechanism.
The analysis provides empirical guidance about how to choose the suitable window size for a dataset.
As shown in Figure \ref{Sensitivity} (a) and (b), when the proportion of the prediction tensor is at 1/64, 1/32, and 1/16, even though the models embody low FLOPs (high efficiency), the accuracies are unsatisfactory. 
When the proportion goes up to 1/8, the accuracies are stable and much better than previous choices. As the growth of proportion rises, there is only a slight increase in accuracy, but the FLOPs grow dramatically. Therefore, to have a better trade-off between accuracy and efficiency, we chose the value of window size around the 1/8 of prediction tensor in our experiments.


\section{Discussions}



In the experiments, our DWT-CapsNet was only combined with ResNet-18. This design is mainly based on two considerations. Firstly, we aim to achieve a more efficient CapsNet-based approach. For this purpose, our primary consideration is to evaluate how our method performs on a small backbone. As shown in the results, DWT-CapsNet has significantly improved the performances of ResNet-18. 
Secondly, when a deeper backbone is employed, such as ResNet-50 and ResNet-100, the number of parameters will increase sharply and the CNN network structure will be unchanged. 
Commonly, the CapsNet uses only a few capsule layers.
This design will make the results of the entire model more biased towards those of deeper CNNs. 
As a result, our method can still work in this situation, but the improvement of DWT-CapsNet will be reduced.

Even though the proposed Pyramid Fusion (PF) is inspired by a Feature Pyramid Network (FPN) \cite{lin2017feature}, our PF, using fewer parameters and lower computational complexity, is more suitable for the HSI classification. 
The conventional FPN will extract the high-level abstract features and then inject the abstract features into the low-level feature maps. The obtained aggregated multi-scale feature maps are helpful for detecting the multi-scale objects on images.
In our case, there is no abstract feature injection process, because we only aim to capture the better part-whole relationships among entities for classification.
As a result, we directly concatenate the prediction tensors on the pyramid, which allows better computational efficiency.

%

In terms of the Partial Connection mechanism, randomly pruning a fixed number of connections between capsule layers is an alternative solution, which is similar to dropout operation enabling an effective ensemble model. 
Compared with our partial connection by using a sliding window method, this solution may perform better due to the powerful ensemble effect.
However, this dropout \cite{JMLR:v15:srivastava14a} operation will introduce another operation for our self-attention mechanism, such as position encoding.
In future work, we will introduce this solution into our method and compare it with our sliding window-based connection.

\section{Conclusions}

In this paper, we aim to propose an efficient and efficient algorithm for HSI classification that is widely used for earth monitoring.
Our studies investigate a novel Discrete Wavelet Transform-based capsule network (DWT-CapsNet) that not only achieves state-of-the-art performance but also recuses the computation demand for its training and inference. 
DWT-CapsNet builds upon a novel attentive DWT downsampling method by integrating a tailored attention mechanism into the 2D-DWT base downsampling method.
Compared to the previous method for learning in the frequency domain,
our design maintains and identifies useful information in different frequency domains for classification.
From the viewpoints of the CapsNet, we propose a novel routing algorithm, where a Pyramid Fusion (PF) structure is able to represent the entities at multiple levels of granularity. 
%
The rich spatial relationship information preserved by the above components allows partial connections among capsule layers so that the number of parameters and computational complexity are reduced dramatically.
Then, we further explore the partial but important connections by using a sliding window and self-attention mechanism in the high-level capsule generation step.
As shown in experimental results, the proposed method shows advantages in both classification performance and computation efficiency.
In the future, as shown in the discussion section, we tend to improve the routing algorithm by randomly pruning connections -- a dropout-based partial connection method.
Additionally, we will deploy the proposed method in real-world and simulated application scenarios.

\bibliography{AAA-sn-bibliography-gzq}

\end{document}